\documentclass[runningheads]{llncs}

\usepackage{graphicx}
\usepackage{amsmath}
\usepackage{makecell}
\usepackage{multirow}
\usepackage{multicol}
\usepackage{booktabs} 
\usepackage{hyperref}
\usepackage{amsfonts,amssymb}
\begin{document}
\title{Neural HD Map Generation from Multiple Vectorized Tiles Locally Produced by Autonomous Vehicles\thanks{This work is supported by the National Natural Science Foundation of China under Grant No. U22A20104. 
For more details about our recent studies, please visit corresponding author's website: \url{https://godfanmiao.github.io/homepage-en/}.
}} 
\titlerunning{Neural HD Map Generation from Vehicle-produced Vectorized Tiles}

\author{\texorpdfstring{Miao Fan$^{\dagger}$,},  Yi Yao, Jianping Zhang, Xiangbo Song, Daihui Wu}
\authorrunning{Fan et al.}

\institute{NavInfo Co., Ltd., Beijing 100094, China\\
\url{https://en.navinfo.com/}\\
\texorpdfstring{$^{\dagger}$}\email{miao.fan@ieee.org}\\
}
\maketitle

\begin{abstract}
High-definition (HD) map is a fundamental component of autonomous driving systems, as it can provide precise environmental information about driving scenes. Recent work on vectorized map generation could produce merely $65\%$ local map elements around the ego-vehicle at runtime by one tour with onboard sensors, leaving a puzzle of how to construct a global HD map projected in the world coordinate system under high-quality standards. To address the issue, we present \textbf{GNMap} as an end-to-end generative neural network to automatically construct HD maps with multiple vectorized tiles which are locally produced by autonomous vehicles through several tours. It leverages a multi-layer and attention-based autoencoder as the shared network, of which parameters are learned from two different tasks (i.e., pretraining and finetuning, respectively) to ensure both the completeness of generated maps and the correctness of element categories. Abundant qualitative evaluations are conducted on a real-world dataset and experimental results show that GNMap can surpass the SOTA method by more than $5\%$ F1 score, reaching the level of industrial usage with a small amount of manual modification. We have already deployed it at Navinfo Co., Ltd., serving as an indispensable software to automatically build HD maps for autonomous driving systems. 

\keywords{HD map \and Autonomous driving \and Vectorized tile \and Multiple tours}
\end{abstract}

\section{Introduction}
\begin{figure}[!t]
\centering
\includegraphics[width=\textwidth]{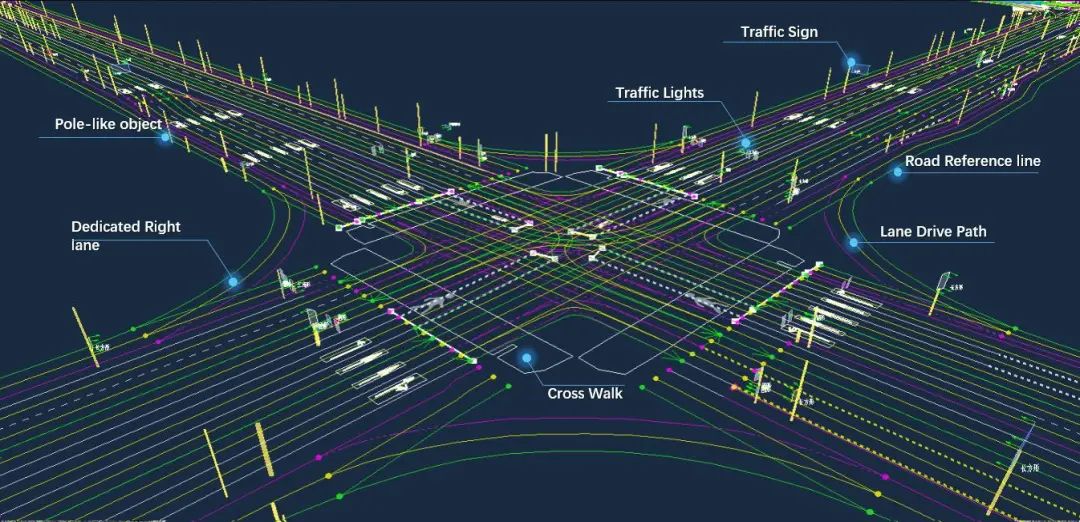}
\vspace{-5mm}
\caption{Illustration of a snapshot of vectorized HD map. It is composed of static map elements, such as pedestrian crossings, lane dividers, road boundaries, etc., which are geometrically discretized into polylines or polygons.} \label{fig1}
\vspace{-5mm}
\end{figure}

High-definition (HD) map~\cite{Elghazaly2023HighDefinitionMC} plays a pivotal role in autonomous driving~\cite{Boubakri2022HighDM,Liu2020HighDM}. Illustrated by Fig.~\ref{fig1}, it provides high-precision vectorized elements (including pedestrian crossings, lane dividers, road boundaries, etc.) about road topologies and traffic rules, which are quite essential for the navigation of self-driving vehicles. Vectorized map elements are geometrically discretized into polylines or polygons, and conventionally produced offline by SLAM-based methods~\cite{Zhang2014LOAMLO,Shan2020LIOSAMTL} with heavy reliance on human labor of annotation, facing both scalability and up-to-date issues.

To address the issues, recent studies~\cite{li2022hdmapnet,liao2022maptr,qiao2023end,shin2023instagram,ding2023pivotnet,liu2023vectormapnet} focus on developing online approaches for vectorized map construction. These methods aim at devising vehicle-mounted models that learn to generate local elements around the ego-vehicle at runtime with onboard sensors such as LiDARs~\cite{Roriz2021AutomotiveLT} and cameras. Learning-based approaches have drawn ever-increasing attention as they can alleviate human efforts to some extent. However, even the SOTA methods~\cite{liao2022maptr,ding2023pivotnet} among them could merely produce $65\%$ vehicle-around map elements by one tour, leaving a puzzle of how to construct a global HD map projected in the world coordinate system under high-quality standards. 

As the first attempt to solve the puzzle, we present \textbf{GNMap} in this paper. It is an end-to-end generative neural network which takes vehicle-produced vectorized tiles through multiple tours as inputs and automatically generates a globalized HD map under the world coordinates as the output.
Specifically, GNMap adopts a multi-layer and attention-based autoencoder as the shared network, of which parameters are learned from two different tasks (i.e., pretraining and finetuning, respectively). At pretraining phase, the shared autoencoder is responsible for completing the masked vectorized tiles. The pretrained parameters are further leveraged as the initial weights for finetuning, which aims at assigning each pixel of map elements to the correct category. In this way, we ensure both the completeness of generated maps and the correctness of element categories. 

Additionally, we build a real-world dataset to conduct qualitative assessments offline. Each instance of the data belongs to a vectorized tile mainly composed of three kinds of map elements, i.e., pedestrian crossings, lane dividers, and road boundaries. Besides that, a tile is passed through multiple tours by autonomous vehicles with a street view for each tour. Ablation studies demonstrate that it is vital to conduct pretraining on GNMap for the sake of achieving the best performance. Experimental results of abundant evaluations also show that it can surpass the SOTA method by more than 5\% F1 score. So far, GNMap has already been deployed at Navinfo Co., Ltd. for industrial usage, serving as an indispensable software to automatically build HD maps of Mainland China for autonomous driving.

\section{Related Work}
\subsection{SLAM-based Methods (Offline)}
HD maps are conventionally annotated manually on LiDAR point clouds of the environment. These point clouds are collected from LiDAR scans of survey vehicles with GPS~\cite{Kaplan1996UnderstandingG} and IMU~\cite{Borodacz2021ReviewAS}. In order to fuse LiDAR scans into an accurate and consistent point cloud, SLAM methods~\cite{Zhang2014LOAMLO,Shan2020LIOSAMTL} are mostly used, and they generally adopt a decoupled pipeline as follows. Pairwise alignment algorithms like ICP~\cite{121791} and NDT~\cite{Biber2003TheND} are firstly employed to match LiDAR data between two nearby timestamps. And for the purpose of constructing a globally consistent map, it is critical to estimate the accurate pose of ego-vehicle by GTSAM~\cite{factor_graphs_for_robot_perception}. Although several machine learning methods~\cite{Mi2021HDMapGenAH} are further devised to extract static map elements such as pedestrian crossings, lane dividers and road boundaries from fused LiDAR point clouds, it is still laborious and costly to maintain a scalable HD map since it requires timely update for autonomous driving.

\subsection{Learning-based Approaches (Online)}
To get rid of offline human efforts, learning-based HD map construction has attracted ever-increasing interests. These approaches~\cite{li2022hdmapnet,liao2022maptr,qiao2023end,shin2023instagram,ding2023pivotnet,liu2023vectormapnet} propose to build local maps at runtime based on surround-view images captured by vehicle-mounted cameras. Specifically, HDMapNet~\cite{li2022hdmapnet} first produces semantic map and then groups pixel-wise semantic segmentation results in the post-processing. VectorMapNet~\cite{liu2023vectormapnet} adopts a two-stage coarse-to-fine framework and utilizes auto-regressive decoder to predict points sequentially, leading to long inference time and the ambiguity about permutation. To alleviate the problem, BeMapNet~\cite{qiao2023end} adopts a unified piece-wise Bezier curve to describe the geometrical shape of map elements. 
InstaGraM~\cite{shin2023instagram} proposes a novel graph modeling for vectorized polylines of map elements that models geometric, semantic and instance-level information as graph representations.
MapTR~\cite{liao2022maptr} uses a fixed number of points to represent a map element, regardless of its shape complexity. PivotNet~\cite{ding2023pivotnet} models map elements through pivot-based representation in a set prediction framework. 
However, even the SOTA methods among them could merely produce $65\%$ vehicle-around map elements by one tour, leaving a puzzle of how to build a global HD map projected under the world coordinates. 

\section{Model}
\subsection{Problem Formulation}
The objective of GNMap is to generate a globalized HD map under the world coordinates from several vehicle-produced tiles. The vehicle-produced tiles are represented by RGB images, and we use $\mathcal{X}$ to denote the set of the images as inputs. As shown by Eq.~\ref{eq1}, GNMap is formulated as $\mathcal{F}(\mathcal{X}; \Theta)$ which learns to fuse the images $\mathcal{X}$ and to generate a globalized HD map as the output denoted by $\mathcal{Y}$:
\begin{equation}
\label{eq1}
\mathcal{Y} = \mathcal{F}(\mathcal{X}; \Theta),
\end{equation}
where $\Theta$ represents the set of best parameters that GNMap needs to explore. 

\subsection{Shared Autoencoder}
To realize $\mathcal{F}(\Theta)$, we devise an autoencoder that is structured into two parts: a neural encoder $E(\mathcal{X}; \theta_{e})$ and a neural decoder $D(\mathcal{Z}; \theta_{d})$. The relationship between the encoder and the decoder is shown by Eq.~\ref{eq2} and Eq.~\ref{eq3}: 
\begin{equation}
\label{eq2}
\mathcal{Z} = E(\mathcal{X}; \theta_{e})
\end{equation}
and
\begin{equation}
\label{eq3}
\mathcal{Y} = D(\mathcal{Z}; \theta_{d}),
\end{equation}
where $E(\mathcal{X}; \theta_{e})$ takes $\mathcal{X}$ as inputs to produce the intermediate feature representation $\mathcal{Z}$ by means of the parameters $\theta_{e}$ of encoder, and $D(\mathcal{Z}; \theta_{d})$ takes intermediate feature $\mathcal{Z}$ as the input to generate the output $\mathcal{Y}$ by means of the parameters $\theta_{d}$ of decoder. Both $\theta_{e}$ and $\theta_{d}$ belong to $\Theta$:
\begin{equation}
\label{eq4}
\Theta = (\theta_{e}, \theta_{d}).
\end{equation}

\begin{figure}[!htp]
\centering
\includegraphics[width=\textwidth]{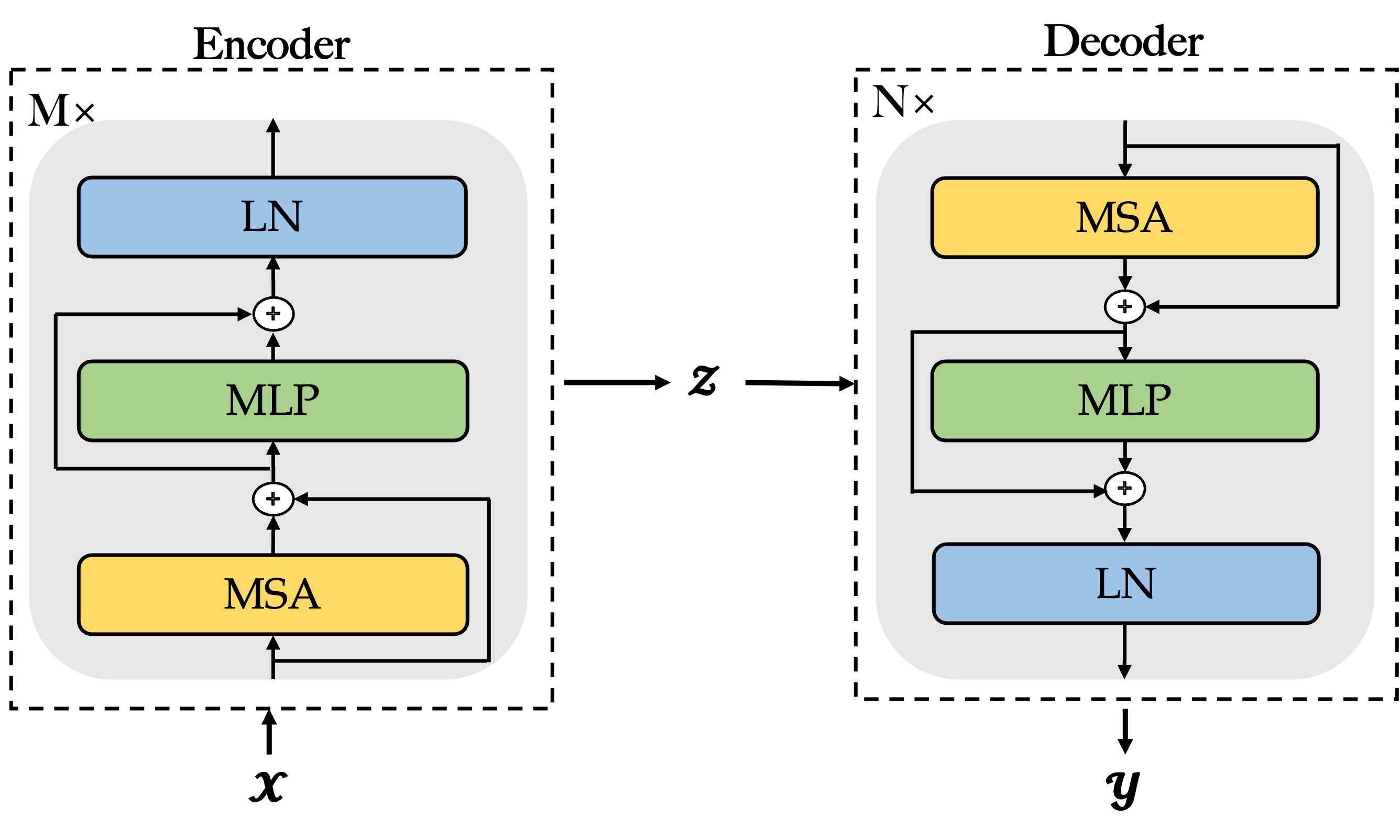}
\vspace{-5mm}
\caption{The architecture of shared autoencoder employed by GNMap. It is a multi-layer generative neural network mainly composed of multi-head self-attention functions.} \label{fig2}
\vspace{-5mm}
\end{figure}
Illustrated by Fig.~\ref{fig2}, both the encoder $E(\mathcal{X}; \theta_e)$ and the decoder $D(\mathcal{Z}; \theta_d)$ are multi-layer networks mainly composed of multi-head self-attention functions. We will elaborate on them in the following paragraphs. 

\paragraph{Encoder:}
$E(\mathcal{X}; \theta_{e})$ is composed of $M$ layer neural blocks with the same structure. Each block includes a multi-head self-attention (MSA~\cite{voita-etal-2019-analyzing}), a multi-layer perceptron (MLP), and a layer normalization (LN) module. Here we use $U_i$ to denote the intermediate output of the block at the $i$-th layer of encoder, and $U_i$ is calculated by Eq. ~\ref{eq5} and Eq.~\ref{eq6}:
\begin{equation}
\label{eq5}
{U}_{i}^{'} = MSA\left ( {U}_{i-1} \right  )+{U}_{i-1},~~~~i\in \{1,2,...,M\}
\end{equation}
and 
\begin{equation}
\label{eq6}
{U}_{i} = LN\left (MLP\left ({U}_{i}^{'} \right )+{U}_{i}^{'}\right),~~i\in \{1,2,...,M\},
\end{equation}
where $\mathcal{X} = U_0$ and $\mathcal{Z} = U_M$.

\paragraph{Decoder:}
$D(\mathcal{Z}; \theta_{d})$ has $N$ stacked blocks with the same structure. Each block is composed of includes a multi-head self-attention (MSA~\cite{voita-etal-2019-analyzing}), a multi-layer perceptron (MLP), and a layer normalization (LN) function as well. If we use $V_j$ to denote the intermediate output of the block at the $j$-th layer of decoder, $V_j$ is calculated by Eq. ~\ref{eq7} and Eq.~\ref{eq8}:
\begin{equation}
\label{eq7}
{V}_{j}^{'} = MSA\left ( {V}_{j-1} \right  )+{V}_{j-1},~~~~j\in \{1,2,...,N\}
\end{equation}
and 
\begin{equation}
\label{eq8}
{V}_{j} = LN\left (MLP\left ({V}_{j}^{'} \right )+{V}_{j}^{'}\right),~~j\in \{1,2,...,N\},
\end{equation}
where $\mathcal{Z} = V_0$ and $\mathcal{Y} = V_N$.

In order to obtain the best parameters of both $\theta_e$ and $\theta_e$, we propose to adopt the "pretraining \& finetuning" manner which divides the training procedure into two phases, corresponding to different tasks and learning objectives. Details about the two phases will be elaborated by Section~\ref{sec:pre} and Section~\ref{sec:finetune}. 

\subsection{Pretraining Phase}
\label{sec:pre}
\begin{figure}
\centering
\includegraphics[width=\textwidth]{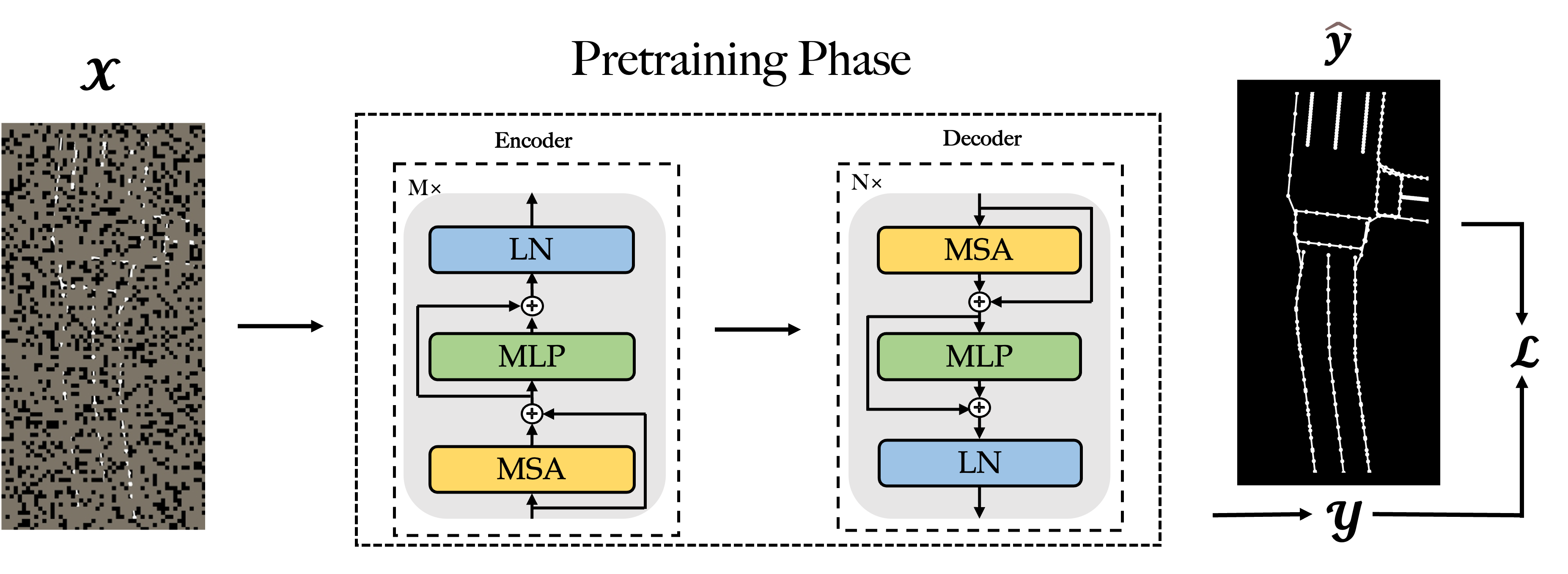}
\vspace{-5mm}
\caption{Illustration of the data processing pipeline at the pretraining phase, where the shared autoencoder is responsible for completing masked (gray-scaled) vectorized tiles.}  
\label{fig3}
\vspace{-5mm}
\end{figure}

At the pretraining phase, the learning objective of the shared autoencoder is to complete masked vectorized tiles, and the pretrained parameters are further leveraged as the initial weights for finetuning. Illustrated by Fig.~\ref{fig3}, we will elaborate pretraining phase from the perspectives of input, output, ground truth, and loss function in the following paragraphs.

\paragraph{Input:}
We split the manually annotated HD maps into multiple vectorized tiles. Each of the vectorized tiles can be transferred into a gray-scaled image denoted by $\mathcal{X} \in \mathbb{R}^{h \times w \times 1}$, where $h$ and $w$ represent the height and the width of the image respectively. In $\mathcal{X}$, each pixel of any may elements is set to 255 and the background's pixel is set to 0.
Then the image is divided into non-overlapping patches with the shape of $k \times l$. As a result, $\frac{h \times w}{k \times l}$ patches (each $p \in \mathbb{R}^{k \times l}$) can be obtained. We sample a subset of patches and mask (i.e., remove) the remaining ones. Our strategy is straightforward: sampling random patches without replacement, following a uniform distribution with a high masking ratio (i.e., the ratio of removed patches). In this way, we have created a task that cannot be easily solved by extrapolation from visible neighboring patches.

\paragraph{Output:} We expect to obtain a completed gray-scale tile as the output through the shared autoencoder which takes the masked patches as inputs. The completed image is denoted by $\mathcal{Y} \in \mathbb{R}^{h \times w \times 1}$, where $h$ and $w$ represent the height and the width of the completed image, respectively. The value of each predicted pixel $y_i$ where $i \in \{1, 2, ..., h \times w\}$ ranges from $0.0$ to $1.0$ since it is scaled by the softmax function. 

\paragraph{Ground Truth:} Correspondingly, the ground-truth image is the unsliced one (i.e., $\mathcal{X}$) used as the input. We denoted it by $\hat{\mathcal{Y}} \in \mathbb{R}^{h \times w \times 1}$ since each pixel of $\hat{\mathcal{Y}}$ is set by either 0 or 1 to indicate whether it belongs to the background or vectorized map elements. 

\paragraph{Loss Function:} We employ the mean squared error (MSE) as the loss function (denoted by $\mathcal{L}$) for pretraining. 
\begin{equation}
\label{pre_loss}
    \mathcal{L} =\frac{1}{h \times w}\displaystyle\sum_{i=1}^{h \times w}{\left ( {y}_{i}-\hat{{y}}_{i}\right )}^{2}.
\end{equation}
As shown by Eq.~\ref{pre_loss}, it measures the overall difference between $\mathcal{Y}$ and $\hat{\mathcal{Y}}$, by calculating the squared errors between the predicted pixels and the ground-truth pixels at the same coordinates. 

\subsection{Finetuning Phase}
\label{sec:finetune}
\begin{figure}[!t]
\centering
\includegraphics[width=\textwidth]{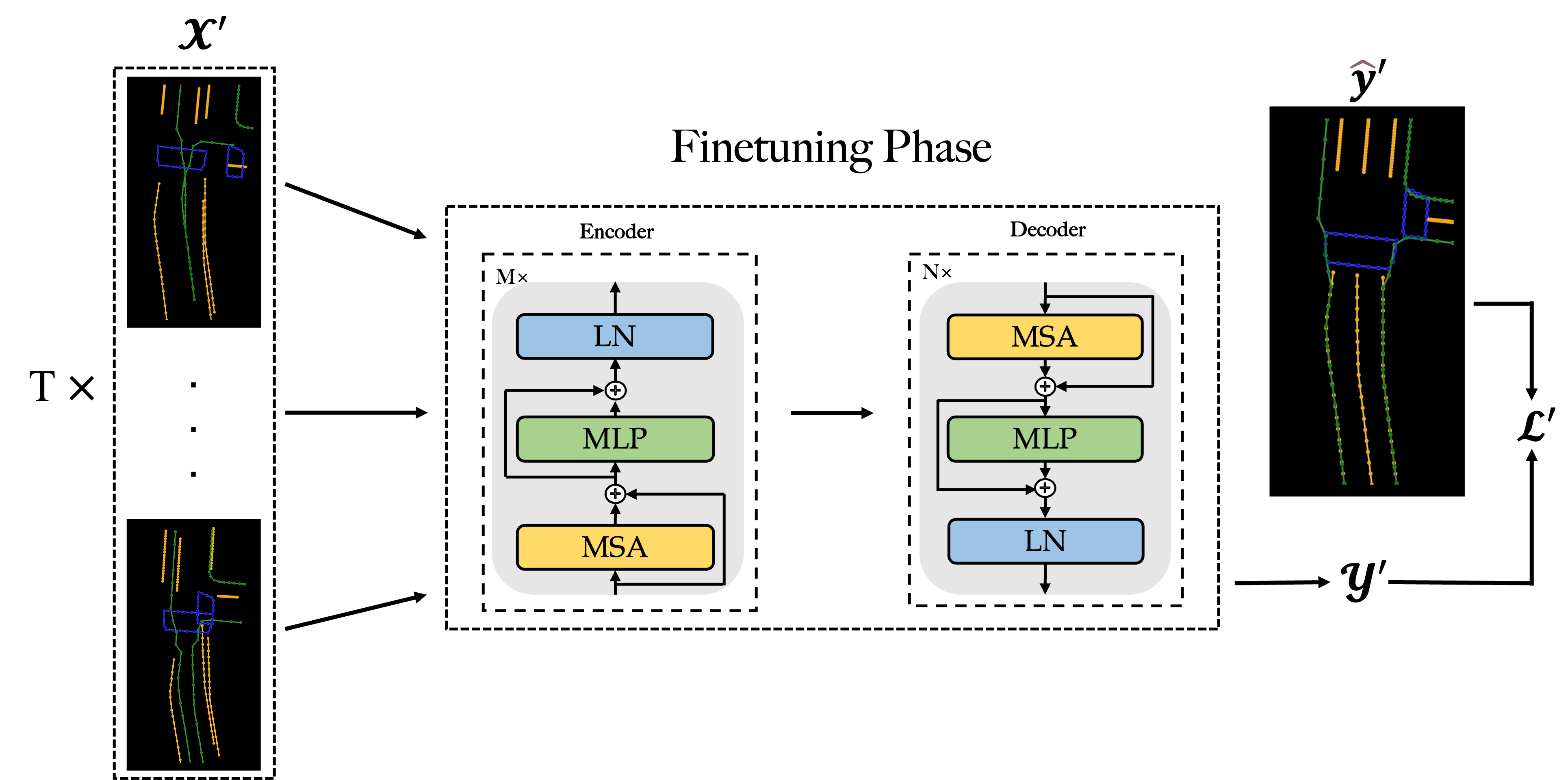}
\vspace{-5mm}
\caption{Illustration of the data processing pipeline at finetuning phase, where the pretrained parameters are leveraged as initial weights of the shared autoencoder.  It aims at assigning each pixel of map elements to the correct category.}  \label{fig4}
\vspace{-5mm}
\end{figure}

At finetuning phase, the learning objective of the shared autoencoder changes to assigning each pixel of the elements of the generated map to the correct category, leveraging the pretrained parameters as initial weights. Illustrated by Fig.~\ref{fig4}, we will elaborate finetuning phase from the perspectives of input, output, ground truth, and loss function in the following paragraphs.

\paragraph{Input:} 
In this work, a tile is passed through $T$ times of tours by autonomous vehicles with a street view for each tour. The original street views collected by the cameras mounted on survey vehicles are usually RGB images and learning-based approaches~\cite{li2022hdmapnet,liao2022maptr,qiao2023end,shin2023instagram,ding2023pivotnet,liu2023vectormapnet} generally transfer them into vectorized images where each pixel belongs to a certain category such as the background or land divider, etc.. 
As a matter of fact, we can obtain $T$ images at the beginning of the finetuning phase. We use a shared CNN network to fetch the features from the $T$ images and concatenate them together as the input of the shared autoencoder. 

\paragraph{Output:} We expect to achieve a fused tile from GNMap as the output at the finetuning phase. The generated image is denoted by $\mathcal{Y} \in \mathbb{R}^{h \times w \times c}$, where $h$ and $w$ represent the height and the width of the image, respectively, and $c$ stands for the kinds of map elements. Each predicted pixel $\textbf{y}_i$ is represented by a $c$-dimensional vector where the value at each dimension ranges from 0.0 to 1.0 to indicate the probability of the predicted category and the sum of all these values is 1.0. 

\paragraph{Ground Truth:} Correspondingly, the ground-truth image is denoted by $\hat{\mathcal{Y}} \in \mathbb{R}^{h \times w \times c}$. In addition, each pixel of $\hat{\mathcal{Y}}$ is denoted by a $c$-dimensional vector where only one of the values is set by 1.0 exclusively indicating that the pixel belongs to a certain category such as the background, pedestrian crossing, or etc.

\paragraph{Loss Function:} We employ the cross-entropy (CE) function as the loss (denoted by $\mathcal{L'}$) of the finetuning phase. 
\begin{equation}
\label{fine_loss}
    \mathcal{L'}=-\frac{1}{h \times w}\displaystyle\sum_{i=1}^{h \times w}\hat{{\textbf{y}'}}_{i} \cdot \log_{}{\left ({\textbf{y}'}_{i} \right )}.
\end{equation}
As shown by ~Eq.~\ref{fine_loss}, it measures the divergence between $\mathcal{Y}$ and $\hat{\mathcal{Y}}$, by summing up the log-likelihood at ground-truth pixels.

\section{Experiments}
\subsection{Dataset and Metrics}
\label{section_dataset}
\begin{table}[!t]
\caption{The statistics of a real-world dataset for the offline assessment of HD map generation from multiple vectorized tiles locally produced by autonomous vehicles in Mainland China. The subsets are separately leveraged for the purpose of model training (abbr. {\it Train}), hyper-parameter tuning (abbr. {\it Valid}), and performance testing (abbr. {\it Test}). Each instance of data belongs to a vectorized \textit{tile} which is mainly composed of several \textit{map elements} (such as pedestrian crossings, lane dividers and road boundaries). Besides that, autonomous vehicles passed through a tile multiple times (\textit{tours}) and collected a \textit{street view} for each tour.}\label{tab1}
\centering
\begin{tabular}{c|c|c|c|c}
\toprule
\textbf{Subset} & \textbf{\#(Tiles)} & \textbf{\#(Map Elements)} & \textbf{Avg. \#(Tours)/Tile} & \textbf{\#(Street Views)}\\
\hline
\hline
Train &  40,000 & 162,493 &  5.2  & 208,207    \\
Valid & 5,000 &  19,928 &  5.0  &  24,982 \\
Test  & 5,000 &  20,061 & 5.1  &  25,564 \\
\bottomrule
\end{tabular}
\vspace{-5mm}
\end{table}
In order to conduct an offline assessment on methods of HD map generation, we build a real-world dataset that contains street views and vectorized tiles produced by autonomous vehicles through multiple tours. We randomly split the dataset into three subsets. As shown by Table~\ref{tab1}, they are separately leveraged for the purpose of model training (abbr. {\it Train}), hyper-parameter tuning (abbr. {\it Valid}), and performance testing (abbr. {\it Test}). Each subset is composed of many exclusive tiles, each of which is passed through multiple \textit{tours} by autonomous vehicles. For each tour, a \textit{street view} is collected and a vectorized tile is produced simultaneously online by vehicle-mounted models. Following up previous work, we mainly focus on three kinds of map elements, including pedestrian crossings (abbr. as ped.), lane dividers (abbr. as div.), and road boundaries (abbr. as bou.).
\begin{figure}[!t]
\centering
\includegraphics[width=\textwidth]{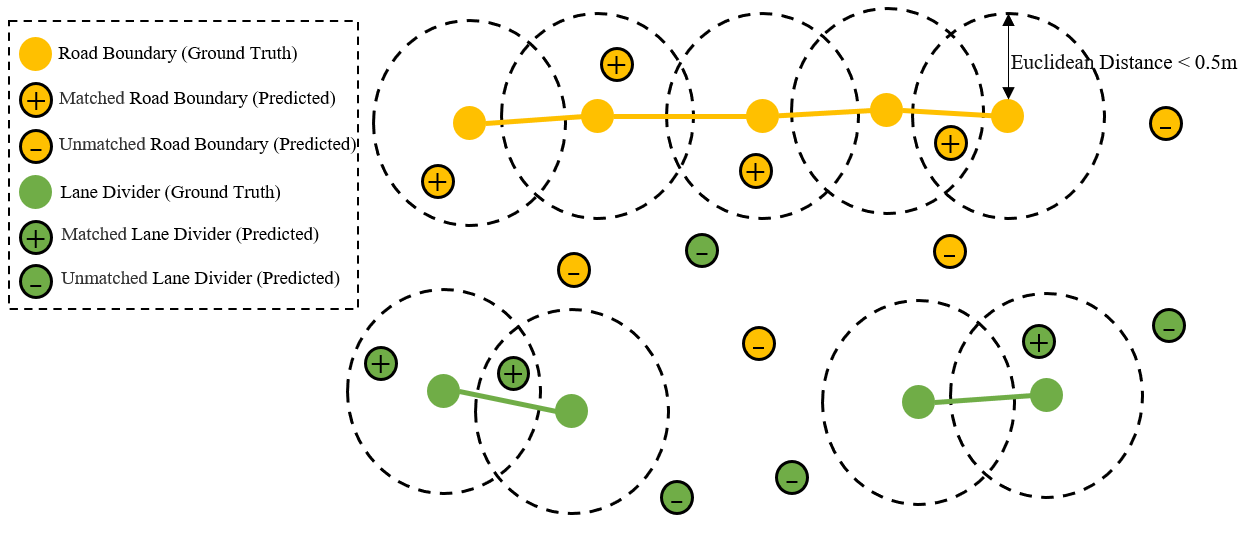}
\vspace{-5mm}
\caption{An example on how to calculate $Precision$ (abbr. as $P$) and $Recall$ (abbr. as $R$). In this case, we have three map elements (two lane dividers and a road boundary). For lane dividers (colored by green), there are 7 predicted points/pixels and 4 ground-truth points/pixels. 3 of 7 are accepted as they locate within 0.5m of the ground-truth pixels. Therefore, $P_{div.} = 3/7$ and $R_{div.} = 3/4$. For the road boundary (colored by yellow), there are 8 predicted points/pixels and 5 ground-truth points/pixels. 4 of 8 are accepted as they locate within 0.5m of the ground-truth pixels. Therefore, $P_{bou.} = 4/8$ and $R_{bou.} = 4/5$.}
\label{fig5}
\vspace{-5mm}
\end{figure}

For each generated tile, we use precision (P) and recall (R) to evaluate the quality of HD map reconstruction at the pixel level in one instance. Illustrated by Fig.~\ref{fig5}, a predicted point is accepted as the positive pixel when it is located near a ground-truth (GT) point within the Euclidean distance of 0.5 meters and they must belong to the same category as well. More importantly, a GT point can only accept one nearest predicted point for evaluation. Assuming that the test set contains $n$ instances, average precision (AP) and average recall (AR) are formulated by Eq.~\ref{eq11} and Eq.~\ref{eq12} as follows,
\begin{equation}
\label{eq11}
{AP}=\frac{1}{n}\displaystyle\sum{P}
\end{equation}
and
\begin{equation}
\label{eq12}
{AR}=\frac{1}{n}\displaystyle\sum{R}
\end{equation}

Then mAP and mAR represent the mean average precision and recall over all categories (i.e., pedestrian crossing, lane divider, and road boundary), which are shown by Eq.~\ref{eq13} and Eq.~\ref{eq14}. 
\begin{equation}
\label{eq13}
mAP = \frac{{AP}_{ped.}+{AP}_{div.}+{AP}_{bou.}}{3}
\end{equation}
\begin{equation}
\label{eq14}
mAR = \frac{{AR}_{ped.}+{AR}_{div.}+{AR}_{bou.}}{3}
\end{equation}
To measure the overall performance of approaches on HD map generation, we adopt F1 score, as shown by Eq.~\ref{eq15}, which calculates the harmonic mean of mAP and mAR.
\begin{equation}
\label{eq15}
{F1} =  \frac{2 \times mAP \times mAR}{mAP+mAR}
\end{equation}

\subsection{Comparison Details}
\begin{table}[!t]
\caption{The experimental results of the offline evaluations on different methods for HD map construction. All the methods are tested by the real-world dataset shown by Table~\ref{tab1} and measured by the metrics mentioned in Section~\ref{section_dataset}}\label{tab2}
\centering
\begin{tabular}{c|c|c|c}
\toprule
\textbf{Method} &  \makecell{\textbf{mAP} \\ \textit{AP}$_{ped.}$ $|$ 
 \textit{AP}$_{div.}$ $|$ \textit{AP}$_{bou.}$} &  \makecell{\textbf{mAR} \\ \textit{AR}$_{ped.}$ $|$ \textit{AR}$_{div.}$ $|$ \textit{AR}$_{bou.}$} & \textbf{F1} \\
\hline
\hline
HDMapNet~\cite{li2022hdmapnet} &  \makecell{45.3 \\ 42.8 $|$ 47.9 $|$ 45.1} & \makecell{44.1 \\ 41.3 $|$ 47.5  $|$ 43.6 }  & 44.7 \\
\hline
VectorMapNet~\cite{liu2023vectormapnet} &  \makecell{62.9 \\  60.4 $|$ 65.3 $|$ 63.1 } & \makecell{61.5 \\   59.2 $|$ 61.8 $|$ 63.4 }  & 62.2 \\
\hline
InstaGraM~\cite{shin2023instagram}& \makecell{53.6 \\ 51.9 $|$ 54.2 $|$ 54.8} & \makecell{62.4 \\ 59.8 $|$ 62.3 $|$  65.1}  & 57.7 \\
\hline
BeMapNet~\cite{qiao2023end} & \makecell{62.3 \\ 60.5 $|$ 61.6 $|$ 64.9} & \makecell{66.1 \\  62.8 $|$ 70.3 $|$ 65.1}  & 64.1 \\
\hline
MapTR~\cite{liao2022maptr} &  \makecell{64.5 \\ 62.8 $|$  65.2 $|$ 65.5} & \makecell{73.2 \\ 71.3 $|$ 73.4 $|$  74.9}  & 68.6 \\
\hline
PivotNet~\cite{ding2023pivotnet} &   \makecell{64.8 \\  63.1 $|$ 66.5 $|$ 64.8 } & \makecell{72.4 \\ 70.3 $|$ 72.8 $|$ 74.1}  & 68.4 \\
\hline
GMM~\cite{reynolds2009gaussian} &  \makecell{63.4 \\ 61.4 $|$ 64.7 $|$ 64.0} & \makecell{63.2 \\ 59.8 $|$   67.6 $|$ 62.3 }  & 63.3 \\
\hline
GNMap (Ours) &  \makecell{72.5 \\ 70.5 $|$ 74.8 $|$ 72.3} & \makecell{75.6 \\ 75.4 $|$  78.1 $|$ 73.3 }  & 74.0 \\
\bottomrule
\end{tabular}
\vspace{-5mm}
\end{table}

We mainly compare GNMap with two groups of approaches. One group contains vehicle-mounted models (including HDMapNet~\cite{li2022hdmapnet}, VectorMapNet~\cite{liu2023vectormapnet}, InstaGraM~\cite{shin2023instagram}, BeMapNet~\cite{qiao2023end}, MapTR~\cite{liao2022maptr}, and PivotNet~\cite{ding2023pivotnet}) which infer vectorized tiles online from real-time street views captured by onsite cameras. The other group represents approaches (i.e., GMM~\cite{reynolds2009gaussian} and our GNMap) on fusing the vehicle-produced tiles to construct a global HD map.
Table~\ref{tab2} reports the experimental results of these two groups of methods for HD map construction. All the approaches are tested by the real-world dataset shown in Table~\ref{tab1} and measured by the metrics mentioned in Section~\ref{section_dataset}. Based on our results, MapTR and PivotNet achieve comparable performance of online map learning through only one tour. Our GNMap outperforms GMM over $10.0\%$ F1 score. 
Even compared with the existing SOTA method of online map learning, GNMap achieves over $5.0\%$ higher F1, demonstrating advanced performance on HD map construction.

\subsection{Ablation Study}
We report ablation experiments in Table~\ref{tab3}, to validate the effectiveness of employing the pretraining phase, and the robustness of using different vehicle-mounted models. We select MapTR~\cite{liao2022maptr} and PivotNet~\cite{ding2023pivotnet}, as the SOTA one-tour vehicle-mounted models, to produce vectorized tiles for GMM~\cite{reynolds2009gaussian} and our GNMap. Experimental results demonstrate that GNMap achieves consistent improvements over GMM regardless of the vehicle-mounted models. Moreover, the pretrained GNMap can provide at least $8.0\%$ higher F1 score than those without pretraining.

\begin{table}[!t]
\caption{Ablation about whether or not to conduct the pretraining phase and to adopt different onsite models that produce vectorized tiles locally.}\label{tab3}

\centering
\begin{tabular}{c|c|c|c}
\toprule
\textbf{Method} &  \makecell{\textbf{mAP} \\ \textit{AP}$_{ped.}$ $|$ 
 \textit{AP}$_{div.}$ $|$ \textit{AP}$_{bou.}$} &  \makecell{\textbf{mAR} \\ \textit{AR}$_{ped.}$ $|$ \textit{AR}$_{div.}$ $|$ \textit{AR}$_{bou.}$} & \textbf{F1} \\
\hline
\hline
GMM (MapTR) &  \makecell{62.5 \\ 61.8 $|$ 63.2 $|$  62.5} & \makecell{66.5 \\ 65.4 $|$ 67.3 $|$ 66.9}  & 64.5 \\
\hline
GNMap (MapTR) w/o Pre. &  \makecell{64.2 \\ 64.3 $|$ 63.6 $|$ 64.8} & \makecell{67.3 \\ 66.3 $|$ 67.4 $|$ 68.3}  & 65.7 \\
\hline
GNMap (MapTR) w/ Pre. &  \makecell{72.7 \\ 70.8 $|$ 74.8 $|$ 72.5  } & \makecell{75.6 \\ 73.3 $|$ 78.1 $|$ 75.4  }  & 74.1 \\

\midrule
GMM (PivotNet) &  \makecell{61.7 \\ 60.9 $|$ 61.5 $|$  62.7} & \makecell{65.6 \\ 64.7 $|$ 66.6 $|$ 65.4}  & 63.6 \\
\hline
GNMap (PivotNet) w/o Pre. &  \makecell{63.8 \\ 62.8 $|$ 63.7 $|$ 64.9} & \makecell{66.5 \\ 65.2 $|$ 66.3 $|$ 67.9}  & 65.1 \\
\hline
GNMap (PivotNet) w/ Pre. & \makecell{72.6 \\ 72.8 $|$ 73.1 $|$ 71.9} & \makecell{75.5 \\  74.2 $|$ 77.3 $|$ 75.1 }  & 74.0 \\
\bottomrule
\end{tabular}
\vspace{-5mm}
\end{table}

\section{Conclusion}
In this paper, we present GNMap as an end-to-end generative framework for HD map construction, which is distinguished from recent studies on producing vectorized tiles locally by autonomous vehicles with onboard sensors such as LiDARs and cameras. GNMap is an essential research to follow up those studies, as it first attempts to fuse multiple vehicle-produced tiles to automatically build a globalized HD map under the world coordinates. To be specific, it adopts a multi-layer autoencoder purely composed of multi-head self-attentions as the shared network, where the parameters are learned from two different tasks (i.e., pretraining and finetuning, respectively) to ensure both the completeness of map generation and the correctness of element categories. Ablation studies demonstrate that it is vital to conduct pretraining on GNMap for the sake of achieving the best performance for industrial usage. And experimental results of abundant evaluations on a real-world dataset show that GNMap can surpass the SOTA method by more than $5\%$ F1 score. So far, it has already been deployed at Navinfo Co., Ltd., serving as an indispensable software to automatically build HD maps of Mainland China for autonomous driving. 

\bibliographystyle{splncs04}
\bibliography{mybib}

\begin{thebibliography}{10}
\providecommand{\url}[1]{\texttt{#1}}
\providecommand{\urlprefix}{URL }
\providecommand{\doi}[1]{https://doi.org/#1}

\bibitem{121791}
Besl, P., McKay, N.D.: A method for registration of 3d shapes. IEEE Transactions on Pattern Analysis and Machine Intelligence  \textbf{14}(2),  239--256 (1992)

\bibitem{Biber2003TheND}
Biber, P., Stra{\ss}er, W.: The normal distributions transform: A new approach to laser scan matching. Proceedings 2003 IEEE/RSJ International Conference on Intelligent Robots and Systems  \textbf{3},  2743--2748 vol.3 (2003)

\bibitem{Borodacz2021ReviewAS}
Borodacz, K., Szczepański, C., Popowski, S.: Review and selection of commercially available imu for a short time inertial navigation. Aircraft Engineering and Aerospace Technology  (2021)

\bibitem{Boubakri2022HighDM}
Boubakri, A., Gammar, S.M., Brahim, M.B., Filali, F.: High definition map update for autonomous and connected vehicles: A survey. 2022 International Wireless Communications and Mobile Computing (IWCMC) pp. 1148--1153 (2022)

\bibitem{factor_graphs_for_robot_perception}
Dellaert, F., Kaess, M.: Factor Graphs for Robot Perception. Foundations and Trends in Robotics, Vol. 6 (2017)

\bibitem{ding2023pivotnet}
Ding, W., Qiao, L., Qiu, X., Zhang, C.: Pivotnet: Vectorized pivot learning for end-to-end hd map construction. In: Proceedings of the IEEE/CVF International Conference on Computer Vision. pp. 3672--3682 (2023)

\bibitem{Elghazaly2023HighDefinitionMC}
Elghazaly, G., Frank, R., Harvey, S., Safko, S.: High-definition maps: Comprehensive survey, challenges, and future perspectives. IEEE Open Journal of Intelligent Transportation Systems  \textbf{4},  527--550 (2023)

\bibitem{Kaplan1996UnderstandingG}
Kaplan, E.D.: Understanding gps : principles and applications (1996)

\bibitem{li2022hdmapnet}
Li, Q., Wang, Y., Wang, Y., Zhao, H.: Hdmapnet: An online hd map construction and evaluation framework. In: 2022 International Conference on Robotics and Automation (ICRA). pp. 4628--4634. IEEE (2022)

\bibitem{liao2022maptr}
Liao, B., Chen, S., Wang, X., Cheng, T., Zhang, Q., Liu, W., Huang, C.: Maptr: Structured modeling and learning for online vectorized hd map construction. In: The Eleventh International Conference on Learning Representations (2023)

\bibitem{Liu2020HighDM}
Liu, R., Wang, J., Zhang, B.: High definition map for automated driving: Overview and analysis. Journal of Navigation  (2020)

\bibitem{liu2023vectormapnet}
Liu, Y., Yuan, T., Wang, Y., Wang, Y., Zhao, H.: Vectormapnet: End-to-end vectorized hd map learning. In: International Conference on Machine Learning. pp. 22352--22369. PMLR (2023)

\bibitem{Mi2021HDMapGenAH}
Mi, L., Zhao, H., Nash, C., Jin, X., Gao, J., Sun, C., Schmid, C., Shavit, N., Chai, Y., Anguelov, D.: Hdmapgen: A hierarchical graph generative model of high definition maps. 2021 IEEE/CVF Conference on Computer Vision and Pattern Recognition (CVPR) pp. 4225--4234 (2021)

\bibitem{qiao2023end}
Qiao, L., Ding, W., Qiu, X., Zhang, C.: End-to-end vectorized hd-map construction with piecewise bezier curve. In: Proceedings of the IEEE/CVF Conference on Computer Vision and Pattern Recognition. pp. 13218--13228 (2023)

\bibitem{reynolds2009gaussian}
Reynolds, D.A., et~al.: Gaussian mixture models. Encyclopedia of biometrics  \textbf{741}(659-663) (2009)

\bibitem{Roriz2021AutomotiveLT}
Roriz, R., Cabral, J., Gomes, T.: Automotive lidar technology: A survey. IEEE Transactions on Intelligent Transportation Systems  \textbf{23},  6282--6297 (2021)

\bibitem{Shan2020LIOSAMTL}
Shan, T., Englot, B., Meyers, D., Wang, W., Ratti, C., Rus, D.: Lio-sam: Tightly-coupled lidar inertial odometry via smoothing and mapping. 2020 IEEE/RSJ International Conference on Intelligent Robots and Systems (IROS) pp. 5135--5142 (2020)

\bibitem{shin2023instagram}
Shin, J., Rameau, F., Jeong, H., Kum, D.: Instagram: Instance-level graph modeling for vectorized hd map learning. arXiv preprint arXiv:2301.04470  (2023)

\bibitem{voita-etal-2019-analyzing}
Voita, E., Talbot, D., Moiseev, F., Sennrich, R., Titov, I.: Analyzing multi-head self-attention: Specialized heads do the heavy lifting, the rest can be pruned. In: Korhonen, A., Traum, D., M{\`a}rquez, L. (eds.) Proceedings of the 57th Annual Meeting of the Association for Computational Linguistics. pp. 5797--5808. Association for Computational Linguistics, Florence, Italy (Jul 2019)

\bibitem{Zhang2014LOAMLO}
Zhang, J., Singh, S.: Loam: Lidar odometry and mapping in real-time. In: Robotics: Science and Systems (2014)

\end{thebibliography}




\end{document}